\documentclass[final]{l4dc2020} 

\title[Learning Hybrid Dynamics and Control]{Hierarchical Decomposition of Nonlinear Dynamics and \\ Control for System Identification and Policy Distillation}

\usepackage{times}

\usepackage{amsfonts}
\usepackage{dsfont}
\usepackage{lscape}

\usepackage{adjustbox}
\usepackage{caption}

\usepackage{pgfplots}
\usepackage{tikz} 

\pgfplotsset{compat=newest}
\usetikzlibrary{decorations.pathmorphing} 
\usetikzlibrary{fit} 
\usetikzlibrary{backgrounds} 
\usetikzlibrary{pgfplots.groupplots}

\DeclareMathOperator*{\argmax}{arg\,max}

\renewcommand{\vec}[1]{\boldsymbol{#1}}
\newcommand{\mat}[1]{\boldsymbol{\mathrm #1}}

\usepackage{natbib}

\makeatletter
\def\set@curr@file#1{\def\@curr@file{#1}} 
\makeatother
\usepackage[load-configurations=version-1]{siunitx} 

\author{%
	\Name{Hany Abdulsamad} $^\dagger$ \Email{abdulsamad@ias.tu-darmstadt.de}\\
	\Name{Jan Peters}  $^{\dagger *}$ \Email{peters@ias.tu-darmstadt.de}\\
	\addr $^\dagger$ Intelligent Autonomous Systems, Technische Universit{\"a}t Darmstadt \\
	$^{*}$ Robot Learning Group, Max Planck Institute for Intelligent Systems%
}

\begin{document}

\maketitle

\begin{abstract}%
	\label{abstract}
	The control of nonlinear dynamical systems remains a major challenge for autonomous agents. Current trends in reinforcement learning (RL) focus on complex representations of dynamics and policies, which have yielded impressive results in solving a variety of hard control tasks. However, this new sophistication and extremely over-parameterized models have come with the cost of an overall reduction in our ability to interpret the resulting policies. In this paper, we take inspiration from the control community and apply the principles of hybrid switching systems in order to break down complex dynamics into simpler components. We exploit the rich representational power of probabilistic graphical models and derive an expectation-maximization (EM) algorithm for learning a sequence model to capture the temporal structure of the data and automatically decompose nonlinear dynamics into stochastic switching linear dynamical systems. Moreover, we show how this framework of switching models enables extracting hierarchies of Markovian and auto-regressive locally linear controllers from nonlinear experts in an imitation learning scenario.%
\end{abstract}


\section{Introduction}%
\label{intro}
The class of nonlinear dynamical systems governs a very wide range of real-world applications, and consequently underpins the most challenging problems of classical control and reinforcement learning \citep{fantoni2002non, kober2013reinforcement}. Recent developments in learning-for-control have pushed towards deploying more complex and highly sophisticated representations, e.g. (deep) neural networks and Gaussian processes, to capture the structure of both dynamics and controllers, leading to overwhelming and unprecedented successes in the domain of RL \citep{mnih2015human, arulkumaran2017brief}. This trend can be observed in both approximate optimal control approaches \citep{deisenroth2011pilco, levine2016end}, and approximate value and policy iteration schemes \citep{schulman2015trust, lillicrap2015continuous, haarnoja2018soft}.

However, before the latest successful revival of neural networks in applications of control and robotics, researchers devised different paradigms for solving difficult control tasks. One interesting concept relied on decomposing nonlinear structures of dynamics and control into simpler local (linear) components, each responsible for an area of the state-action space. This decomposition is done with the aim of preserving interpretability and favorable mathematical properties studied over decades in classical control theory, such as local linear-quadratic assumptions \citep{liberzon2011calculus}. Parallels to this paradigm can be found in the control literature under the labels of \textit{Hybrid Systems} or \textit{Switched Models} \citep{liberzon2003switching, haddad2006impulsive, goebel2012hybrid, borrelli2017predictive}, while in the machine and reinforcement learning communities the terminology of \textit{Switching Dynamical Systems} (SDS) and \textit{Switching State-Space Models} (SSM) is more widely adopted \citep{ghahramani2000variational, beal2003variational, fox2009bayesian, linderman2017recurrent}.

Starting from this paradigm, we present in this work a view of data-driven automatic system identification and learning of composite control from the perspective of hybrid systems and switching linear dynamics. We are motivated by recent in-depth analysis and implications of using piecewise linear (PWL) activation functions such as rectified linear units (ReLU) \citep{montufar2014number, arora2016understanding, pan2016expressiveness, serra2017bounding, petersen2018optimal}, which show, that such representations effectively divide the input space into linear sub-regions.

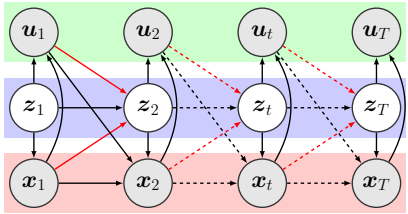
\begin{figure}[t]
	\begin{minipage}{0.45\textwidth}
		\vspace{-0.35cm}
		\hspace{-0.65cm}
		\resizebox{\textwidth}{!}{\centering
\tikzstyle{input}=[circle,
very thick,
minimum size=1.5cm,
draw=black!80,
fill=white!20
]

\tikzstyle{output}=[circle,
very thick,
minimum size=1.5cm,
draw=black!80,
fill=gray!20
]

\tikzstyle{background1}=[rectangle,
fill=green!20,
inner sep=0.15cm,
rounded corners=0mm,
]

\tikzstyle{background2}=[rectangle,
fill=blue!20,
inner sep=0.15cm,
rounded corners=0mm,
]

\tikzstyle{background3}=[rectangle,
fill=red!20,
inner sep=0.15cm,
rounded corners=0mm,
]

\begin{tikzpicture}[>=latex,text height=1.5ex,text depth=0.25ex, scale=0.1]
\matrix[row sep=0.8cm,column sep=1.0cm] {
	& 
	\node (u_1) [output]{\huge $\vec{u}_1$}; & &
	\node (u_2) [output]{\huge $\vec{u}_2$}; & &
	\node (u_t) [output]{\huge $\vec{u}_t$}; & &
	\node (u_T) [output]{\huge $\vec{u}_T$}; & &
	\\ & 
	\node (z_1) [input]{\huge $\vec{z}_1$}; & &
	\node (z_2) [input]{\huge $\vec{z}_2$}; & &
	\node (z_t) [input]{\huge $\vec{z}_t$}; & &
	\node (z_T) [input]{\huge $\vec{z}_T$}; & &
	\\ & 
	\node (x_1) [output]{\huge $\vec{x}_1$}; & &
	\node (x_2) [output]{\huge $\vec{x}_2$}; & &
	\node (x_t) [output]{\huge $\vec{x}_t$}; & &
	\node (x_T) [output]{\huge $\vec{x}_T$}; & &
	\\
};
\path[->]
(z_1) edge[very thick] (z_2)
(z_2) edge[very thick, dashed] (z_t)
(z_t) edge[very thick, dashed] (z_T)

(x_1) edge[very thick] (x_2)
(x_2) edge[very thick, dashed] (x_t)
(x_t) edge[very thick, dashed] (x_T)

(z_1) edge[very thick] (u_1)
(z_2) edge[very thick] (u_2)
(z_t) edge[very thick] (u_t)
(z_T) edge[very thick] (u_T)


(x_1) edge[bend right=30, very thick] (u_1)
(x_2) edge[bend right=30, very thick] (u_2)
(x_t) edge[bend right=30, very thick] (u_t)
(x_T) edge[bend right=30, very thick] (u_T)

(x_1) edge[red, very thick] (z_2)
(x_2) edge[red, very thick, dashed] (z_t)
(x_t) edge[red, very thick, dashed] (z_T)

(u_1) edge[red, very thick] (z_2)
(u_1) edge[black, very thick] (x_2)
(u_2) edge[red, very thick, dashed] (z_t)
(u_2) edge[black, very thick, dashed] (x_t)
(u_t) edge[red, very thick, dashed] (z_T)
(u_t) edge[black, very thick, dashed] (x_T)

(z_1) edge[very thick] (x_1)
(z_2) edge[very thick] (x_2)
(z_t) edge[very thick] (x_t)
(z_T) edge[very thick] (x_T)
;

\begin{pgfonlayer}{background}
	\node [background1, fit=(u_1) (u_T), label=left:] {};
	\node [background2,	fit=(z_1) (z_T), label=left:] {};
	\node [background3, fit=(x_1) (x_T), label=left:] {};
	\node [fit=(x_1) (x_T),	label=left:] {};
\end{pgfonlayer}

\end{tikzpicture}}
	\end{minipage}
	\hspace{-1.05cm}
	\begin{minipage}{0.6\textwidth}
		\caption{Graphical model of recurrent auto-regressive hidden Markov models (rAR-HMM) as presented in \citep{linderman2017recurrent}, extended to support hybrid controls. rAR-HMMs are hybrid dynamic Bayesian networks (HDBN) that explicitly allow the discrete state $z$ to depend on the continuous variables $\vec{x}$ and $\vec{u}$, as highlighted in red.}
		\label{fig:rarhmm}
	\end{minipage}
	\vspace{-.55cm}
\end{figure}

Furthermore, our interest in hybrid systems is motivated by favorable properties inherent in such models. Besides the advantages of tractably acquiring generative models through powerful unsupervised Bayesian inference techniques, hybrid systems allow modeling of discrete events and hard nonlinearities. Moreover, they include built-in time recurrency, that enables them to capture correlations over extended time horizons, an important property when modeling physical systems governed by discrete-time ordinary differential equations. Finally, given the local scope of these systems, they are suitable for modeling phenomena with state-dependent noise.

In the following, we introduce the notation of fully observable stochastic switching linear systems and derive an expectation-maximization algorithm for inferring the parameters of a probabilistic graphical model of hybrid closed-loop dynamics. We use this inference technique to simultaneously learn the system dynamics and decompose nonlinear controllers from state-of-the-art algorithms into simpler, local Markovian and auto-regressive policies, thus dramatically reducing the complexity of dynamics and policy models. We benchmark the predictive power of the learned dynamics against a set of other commonly used representations on simulated classical control tasks.

\section{Hybrid Systems as Dynamic Bayesian Networks}
\label{sec:prelim}
We focus on \textit{Recurrent Auto-Regressive Hidden Markov Models} (rAR-HMM), a special case of Switching Linear Dynamical Systems (SLDS) as presented in \citep{linderman2017recurrent} and implicitly discussed in \citep{barber2006expectation}. We extend them to support exogenous and endogenous inputs in order to simulate the open- and closed-loop behaviors of driven dynamics. Figure~\ref{fig:rarhmm} depicts the extended rAR-HMM as a hybrid dynamic Bayesian network (HDBN). The resulting graphical model is, in our opinion, most suitable for modeling dynamical systems and closely resembles \textit{Piecewise Auto-Regressive Models} (PWARX) \citep{paoletti2007}, widely used in the control literature.

\begin{figure}[t!]
	\hyphenpenalty=10000
	\exhyphenpenalty=10000
	\begin{minipage}[c]{0.6\columnwidth}
		\begin{minipage}[t]{0.15\columnwidth}%
			\input{figures/hybrid_dyn_0.tex}%
		\end{minipage}\hspace*{2.cm}%
		\begin{minipage}[t]{0.15\columnwidth}%
			\input{figures/hybrid_dyn_1.tex} %
		\end{minipage}\hspace*{1.5cm}%
		\begin{minipage}[t]{0.15\columnwidth}%
			\input{figures/hybrid_dyn_2.tex}%
		\end{minipage}
		\begin{minipage}[t]{0.15\columnwidth}%
\begin{tikzpicture}

\begin{axis}[
width=4.cm,
height=4.cm,
xmin=-10, xmax=10,
ymin=-10, ymax=10,
ticks=none,
x grid style={white!69.01960784313725!black},
y grid style={white!69.01960784313725!black},
y label style={yshift=0.1em},
ylabel=${x}_{2}$,
x label style={yshift=0em},
xlabel=${x}_{1}$,
]
\addplot graphics [includegraphics cmd=\pgfimage,xmin=-10, xmax=10, ymin=-10, ymax=10] {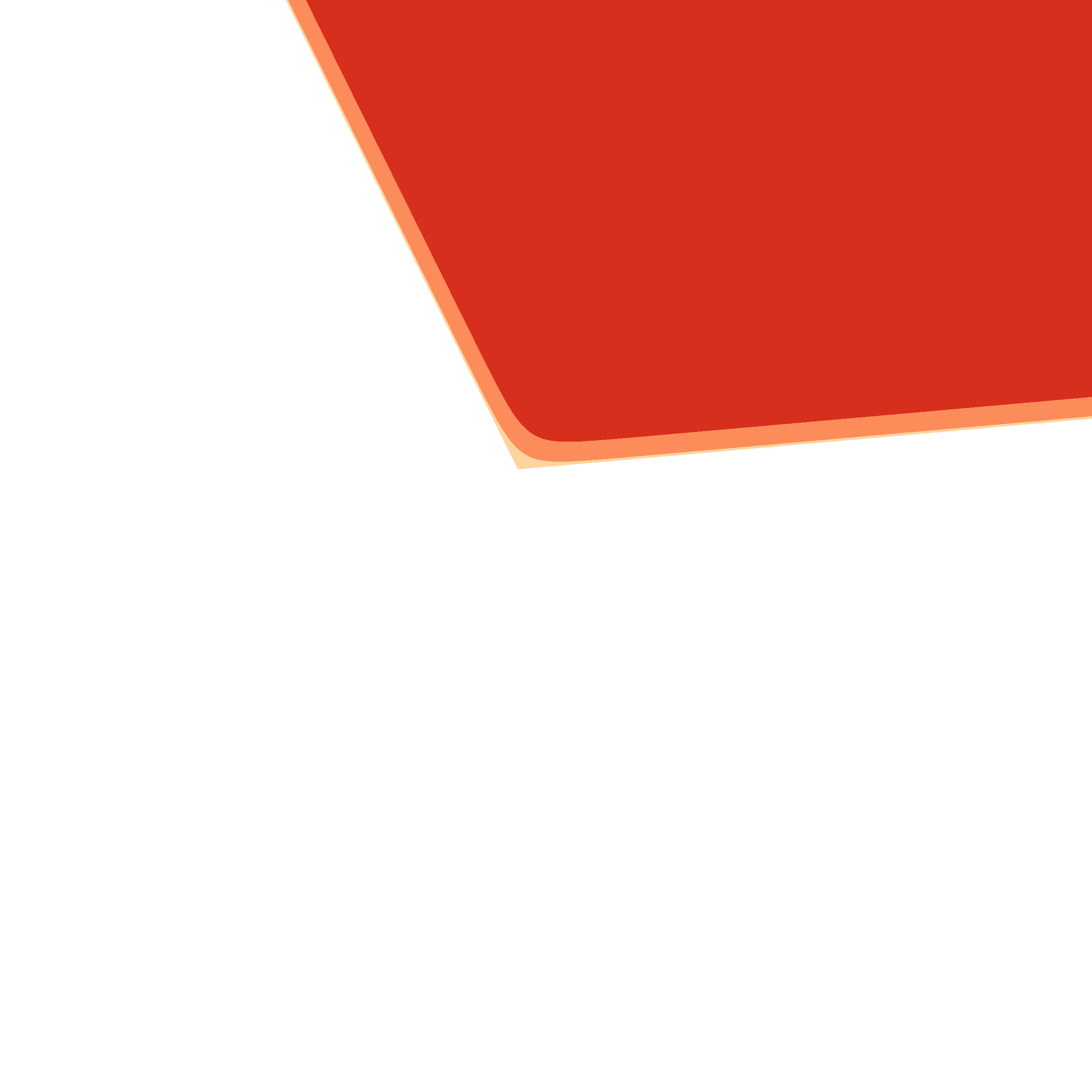};
\addplot graphics [includegraphics cmd=\pgfimage,xmin=-10, xmax=10, ymin=-10, ymax=10] {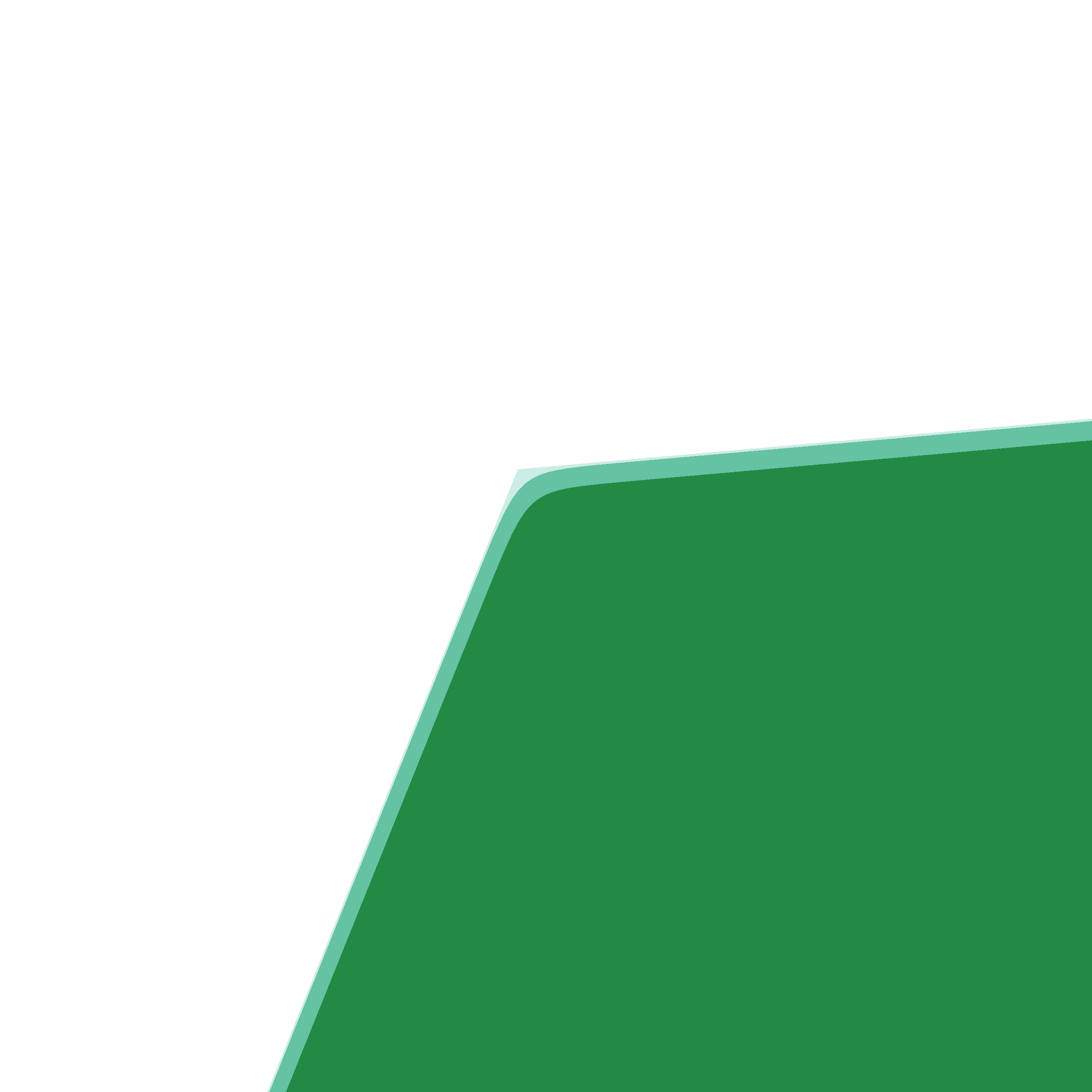};
\addplot graphics [includegraphics cmd=\pgfimage,xmin=-10, xmax=10, ymin=-10, ymax=10] {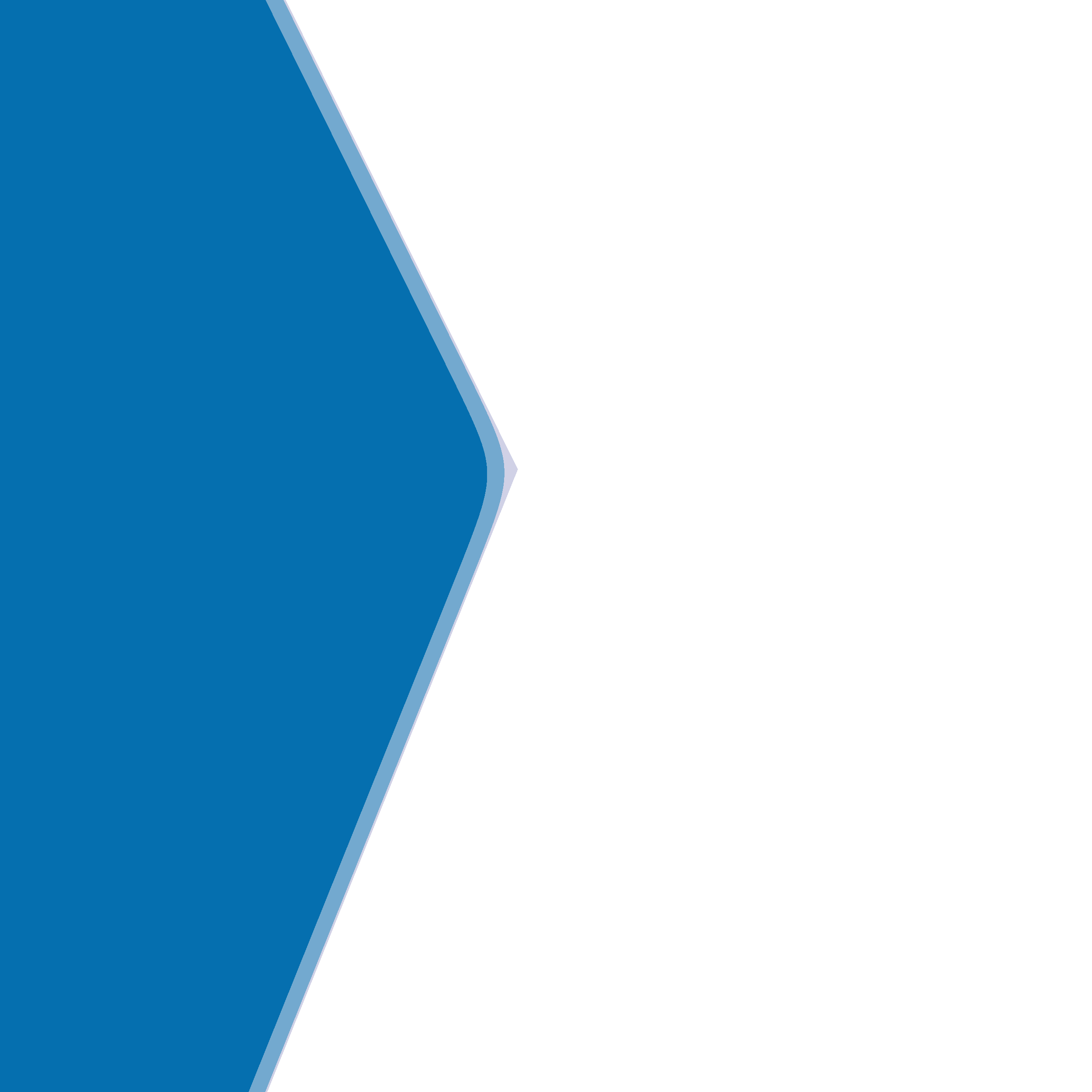};
\end{axis}

\end{tikzpicture}%
		\end{minipage}\hspace*{2.cm}%
		\begin{minipage}[t]{0.15\columnwidth}%
\begin{tikzpicture}

\begin{axis}[
width=4.cm,
height=4.cm,
xmin=-10, xmax=10,
ymin=-10, ymax=10,
ticks=none,
x grid style={white!69.01960784313725!black},
y grid style={white!69.01960784313725!black},
y label style={yshift=-1.5em},
x label style={yshift=0em},
xlabel=${x}_{1}$,
]
\addplot graphics [includegraphics cmd=\pgfimage,xmin=-10, xmax=10, ymin=-10, ymax=10] {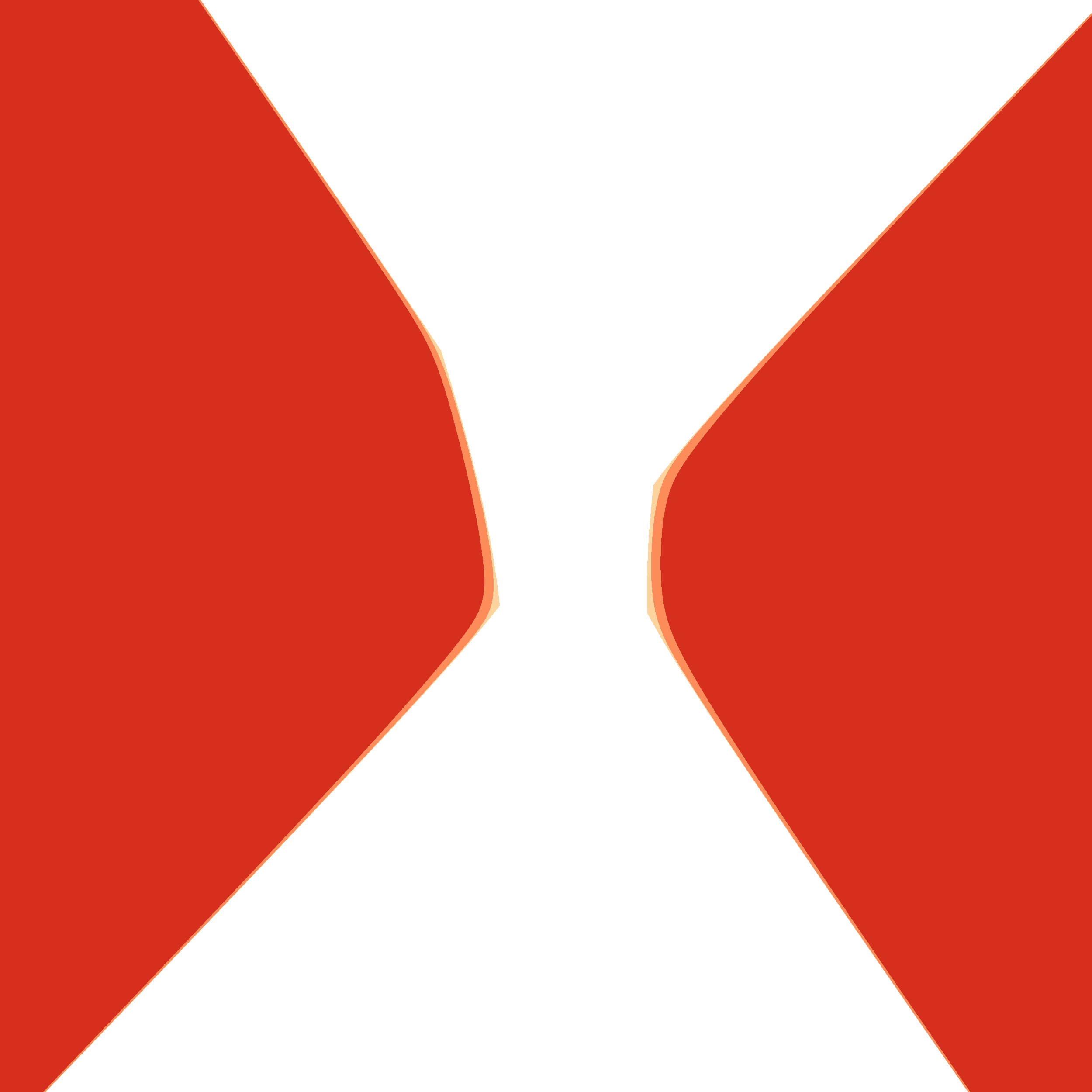};
\addplot graphics [includegraphics cmd=\pgfimage,xmin=-10, xmax=10, ymin=-10, ymax=10] {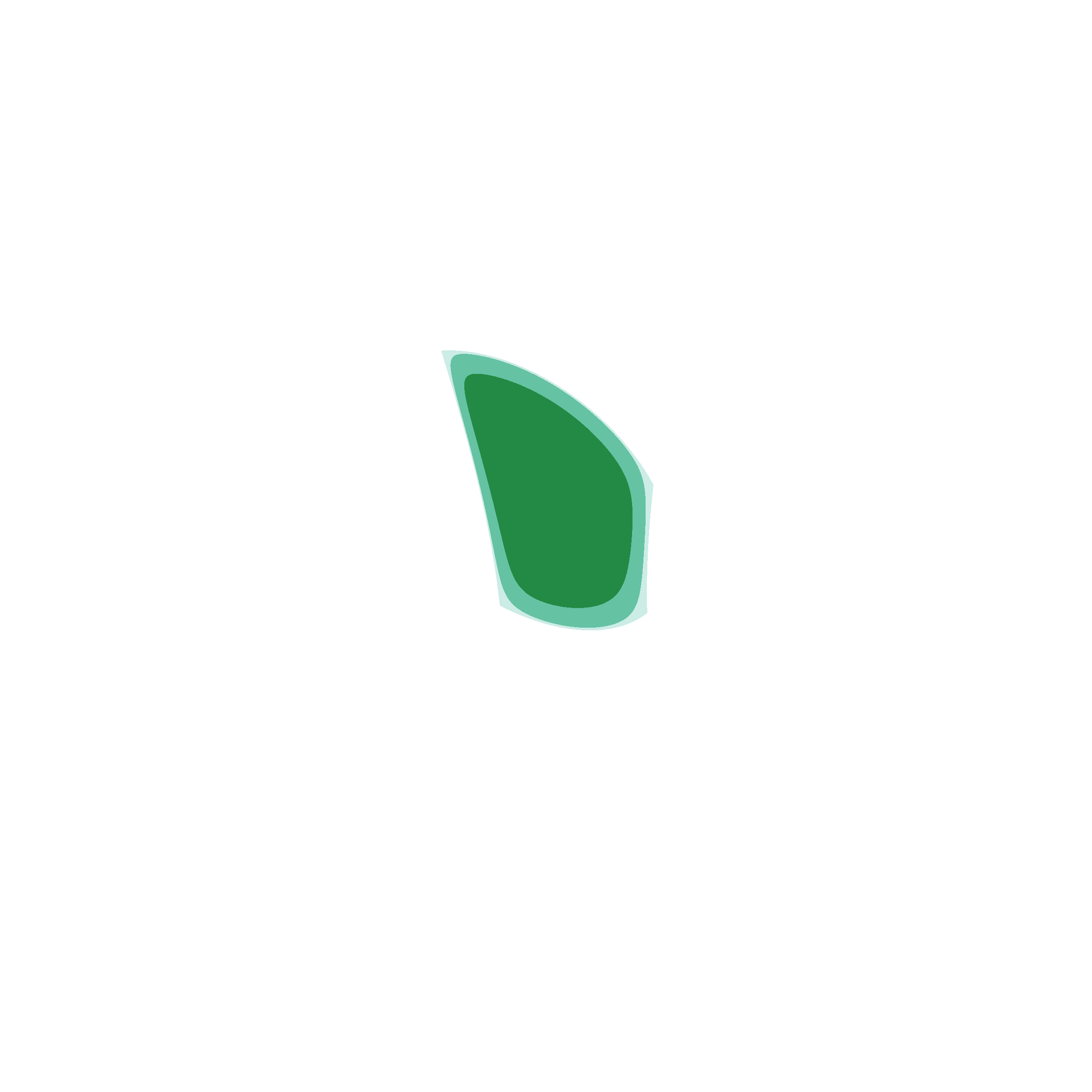};
\addplot graphics [includegraphics cmd=\pgfimage,xmin=-10, xmax=10, ymin=-10, ymax=10] {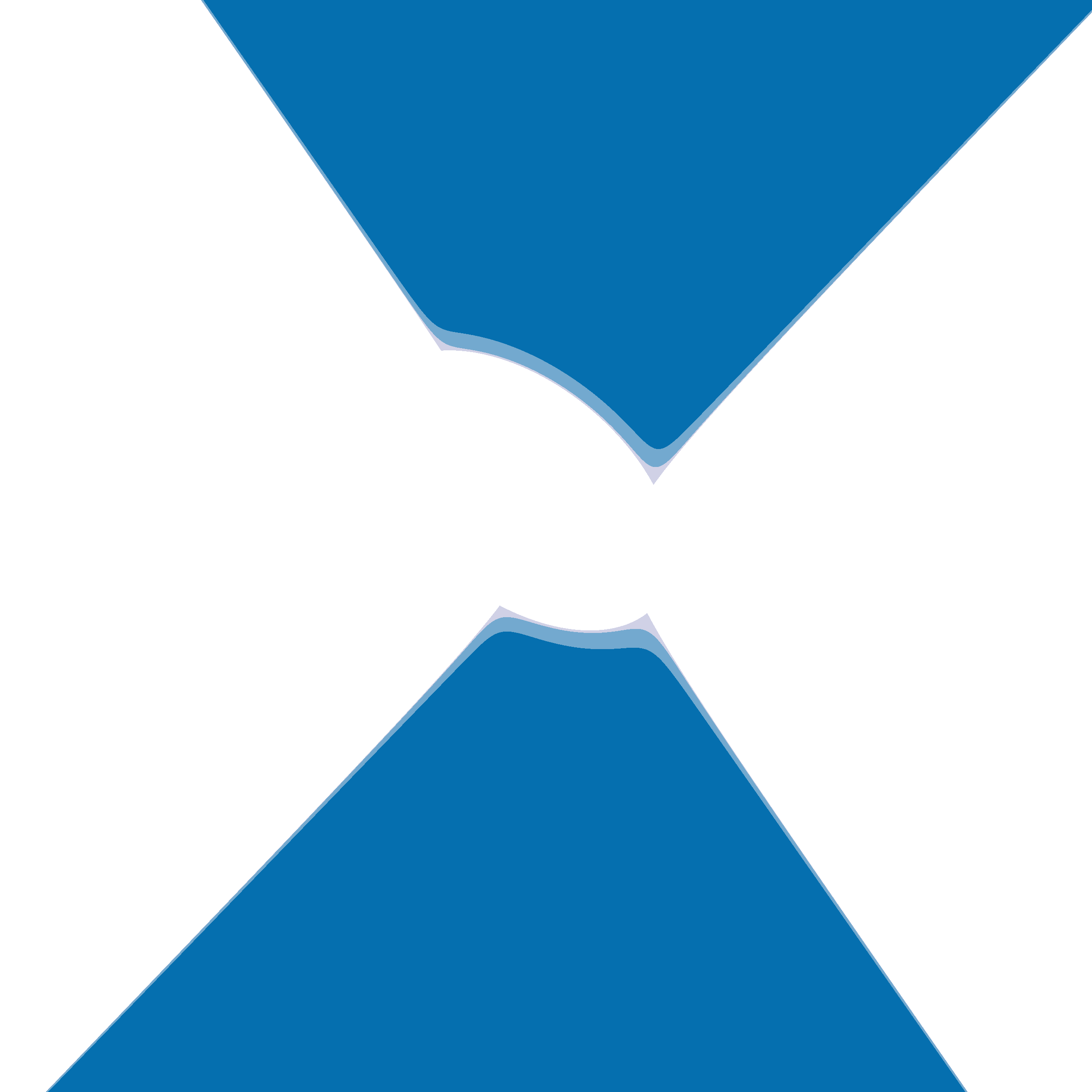};
\end{axis}

\end{tikzpicture}%
		\end{minipage}\hspace*{1.5cm}%
		\begin{minipage}[t]{0.15\columnwidth}%
\begin{tikzpicture}

\begin{axis}[
width=4.cm,
height=4.cm,
xmin=-10, xmax=10,
ymin=-10, ymax=10,
ticks=none,
x grid style={white!69.01960784313725!black},
y grid style={white!69.01960784313725!black},
y label style={yshift=-1.5em},
x label style={yshift=0em},
xlabel=${x}_{1}$,
]
\addplot graphics [includegraphics cmd=\pgfimage,xmin=-10, xmax=10, ymin=-10, ymax=10] {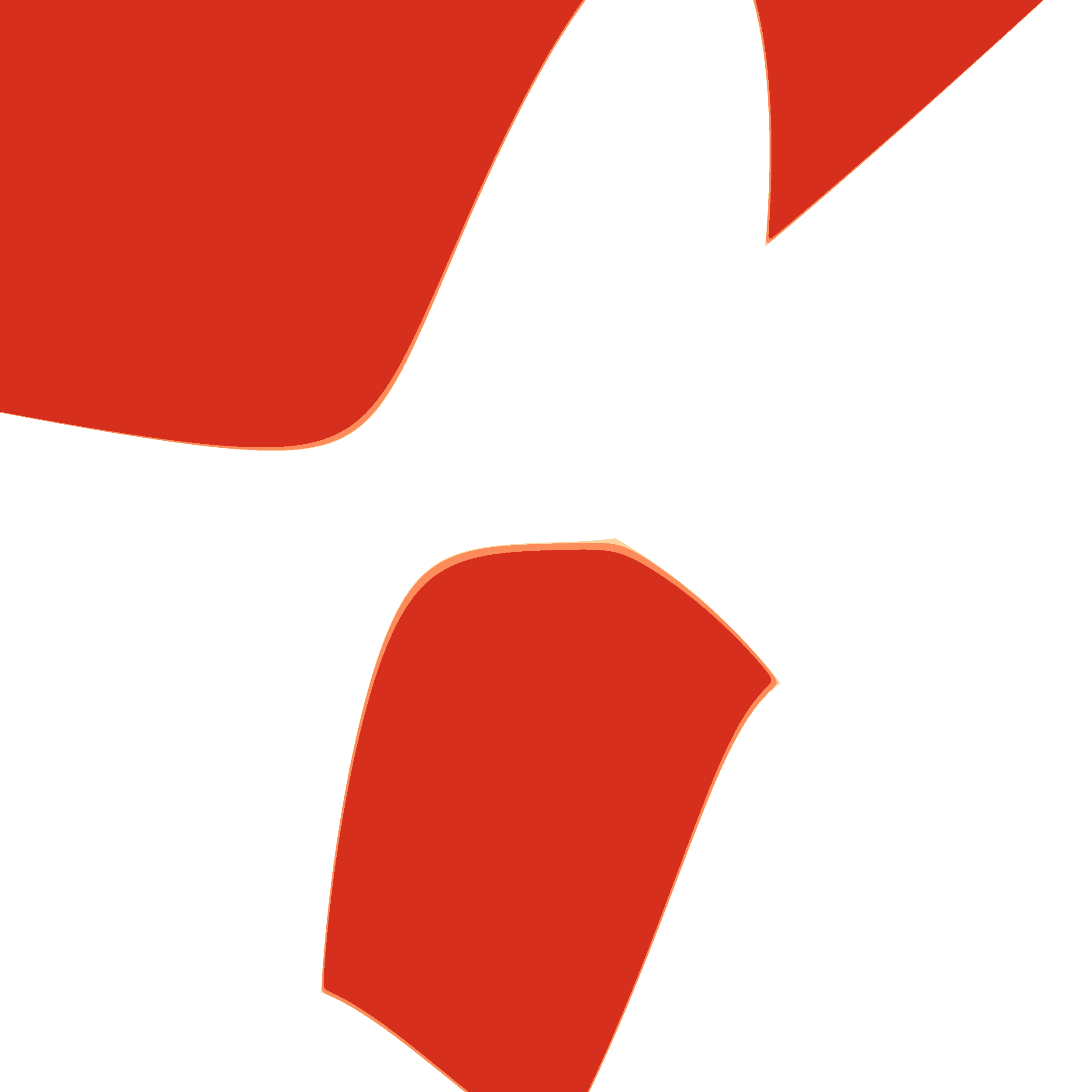};
\addplot graphics [includegraphics cmd=\pgfimage,xmin=-10, xmax=10, ymin=-10, ymax=10] {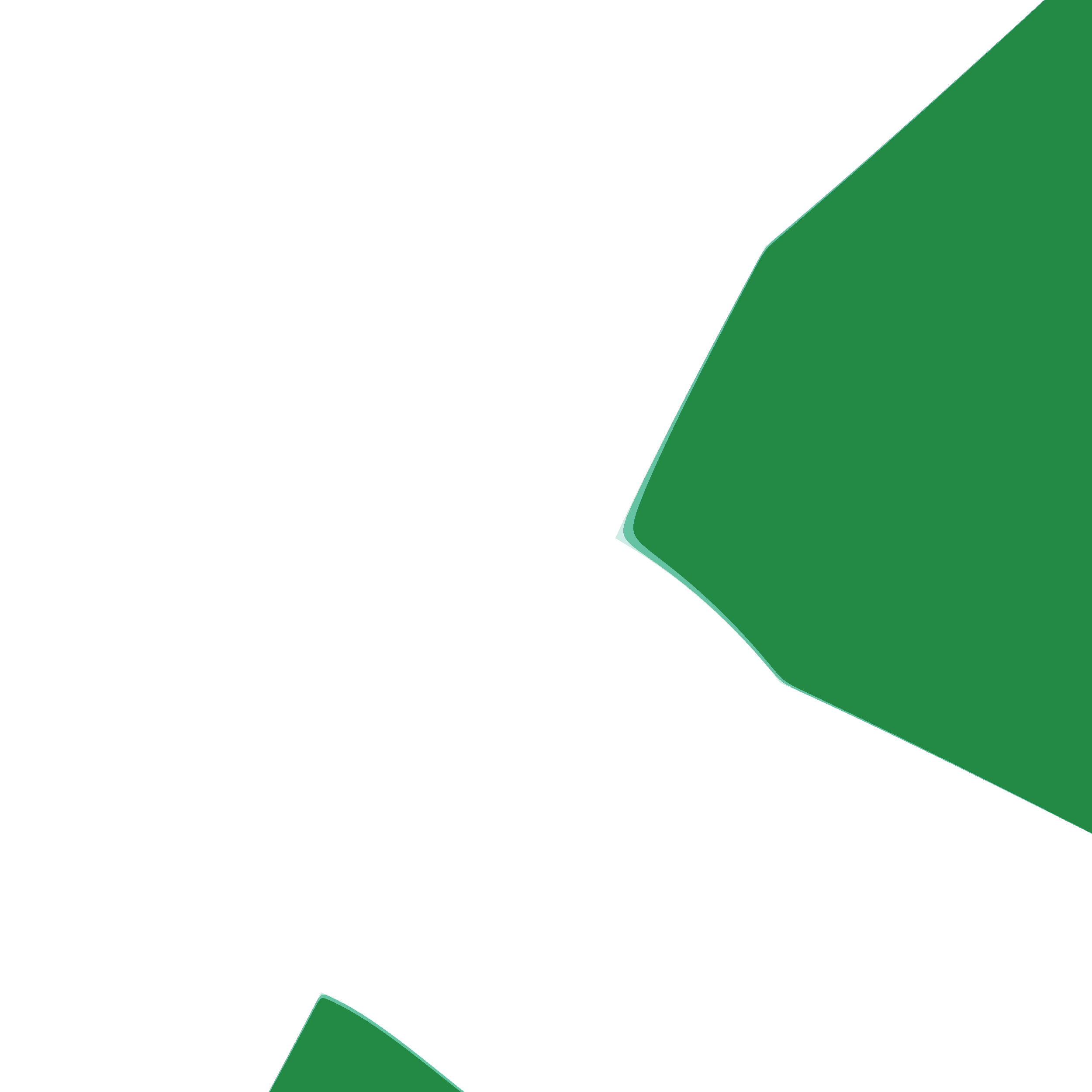};
\addplot graphics [includegraphics cmd=\pgfimage,xmin=-10, xmax=10, ymin=-10, ymax=10] {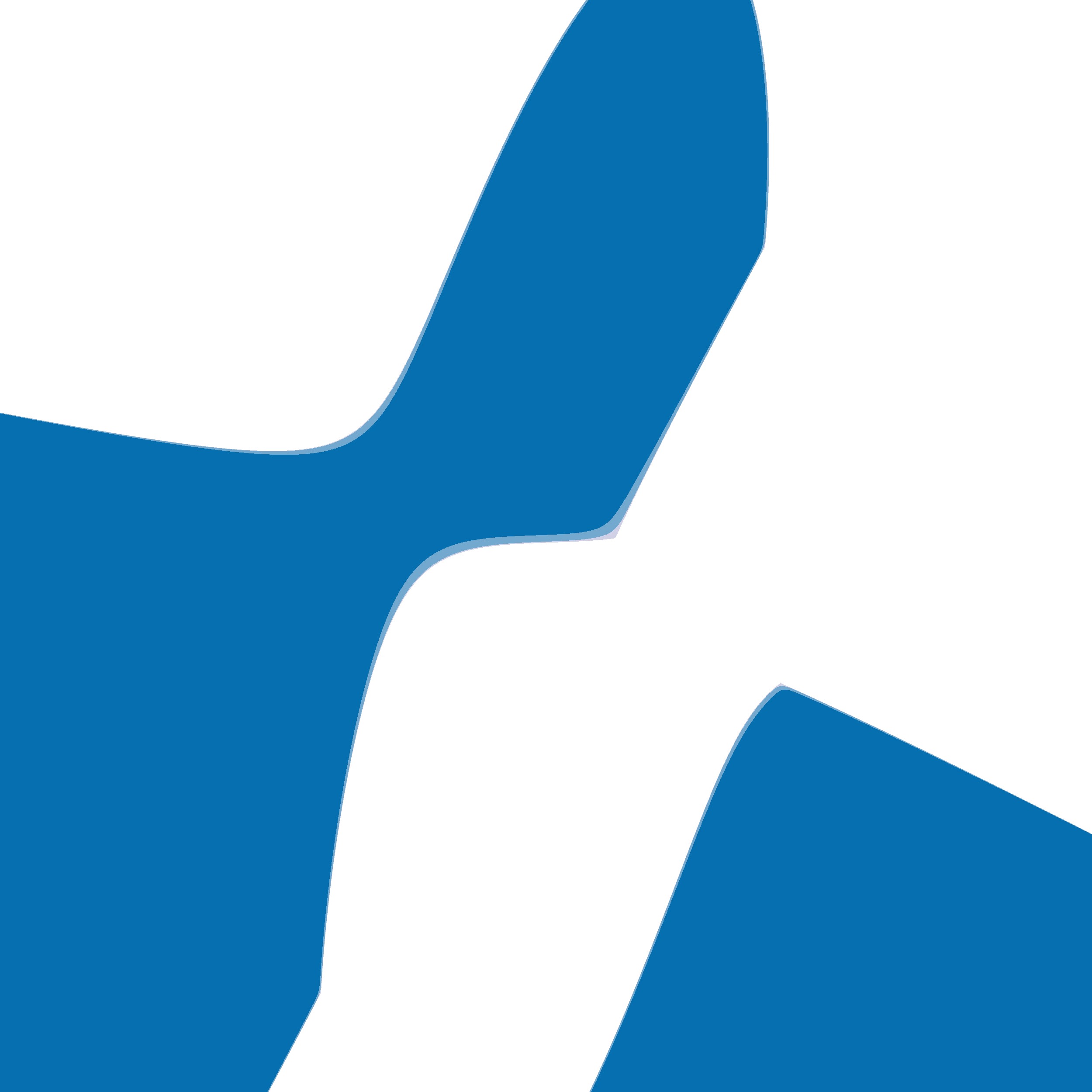};
\end{axis}

\end{tikzpicture}%
		\end{minipage}
	\end{minipage}
	\begin{minipage}[c]{0.35\columnwidth}
		\vspace{-0.1cm}
		\caption{A hybrid system with $K=3$ local linear regimes. The top figures depict the mean unforced continuous transition dynamics in the phase space. The lower figures show the probability of switching, with corresponding color, as a function of the state. We show different decision boundary models: linear (left), quadratic (middle) and third order polynomial (right) functions.}
		\label{fig:rarhmm_dyn}
	\end{minipage}
	\vspace{-0.5cm}
\end{figure}

An rAR-HMM with $K$ regions models the trajectory of a hybrid system as follows. The initial continuous state $\vec{x}_{1}$ and continuous action $\vec{u}_{1}$ are are drawn from a conditional Gaussian distributions, while the initial region indicator variable $z_{1}$ is modeled by a categorical distribution
\begin{equation}
	z_{1} \sim \text{Cat}(K, \vec{\pi}), \quad \vec{x}_{1}|z_{1} \sim \mathcal{N}(\vec{\mu}_{z_{1}}, \mat{\Omega}_{z_{1}}), \quad \vec{u}_{1}|z_{1},\vec{x}_{1} \sim \mathcal{N}(\mat{K}_{z_{1}}\vec{x}_{1},\mat{\Sigma}_{z_{1}}).
\end{equation}
The transition of the continuous state $\vec{x}_{t+1}$ and actions $\vec{u}_{t}$ are modeled by linear-Gaussian dynamics
\begin{alignat}{3}
	 & \vec{x}_{t+1} &  & = \mat{A}_{z_{t+1}}\vec{x}_{t} + \mat{B}_{z_{t+1}}\vec{u}_{t} + \vec{c}_{z_{t+1}} + \vec{\lambda}_{z_{t+1}}, \qquad &  & \vec{\lambda}_{z}\sim\mathcal{N}(\vec{0},\mat{\Lambda}_{z}),  \label{eq:rarhmm_linear}				\\
	 & \vec{u}_{t}   &  & = \mat{K}_{z_{t}} \vec{x}_{t} + \vec{\sigma}_{z_{t}},                                                               &  & \vec{\sigma}_{z}\sim\mathcal{N}(\vec{0},\mat{\Sigma}_{z}),                  
\end{alignat}
where $\left \{\mat{A}, \mat{B}, \vec{c}, \mat{K}, \mat{\Sigma}, \mat{\Lambda} \right \}$ are matrices and vectors of appropriate dimensions with respect to $\vec{x}$ and $\vec{u}$, $\forall k\in[1,K]$. The transition probability $p(z_{t+1}|z_{t}, \vec{x}_{t}, \vec{u}_{t})$, marking the next active dynamical regime, is governed by $K$ categorical distributions, whose probabilities are determined by $K$ state-action dependent generalized linear models (GLM) \citep{gelman2013bayesian} with logistic link functions
\begin{align}
	\psi_{ij} = p(z_{t+1}=i|z_{t}=j,\vec{x}_{t},\vec{u}_{t}) = \frac{\exp\left(f(\vec{x}_{t},\vec{u}_{t};\vec{\omega}_{ij})\right)}{\sum_{k}\exp\left(f(\vec{x}_{t},\vec{u}_{t};\vec{\omega}_{kj})\right)}, \label{eq:rarhmm_lgstc}
\end{align}
where $f$ may contain any type of features of $\vec{x}$ and $\vec{u}$ and is parameterized by $\vec{\omega}_{ij}$, a specific vector for each transition $j \rightarrow i~ \forall {i,j} \in [1, K]$. Figure~\ref{fig:rarhmm_dyn} depicts random realizations of different transition functions that lead to a variety of decompositions of the state space.

This representation of switching dynamics has a major advantage in comparison to other suggested switching models \citep{fox2009nonparametric}, as it allows for modeling the discrete transition dynamics as a function of both the continuous state $\vec{x}$ and action $\vec{u}$, thus coupling continuous and discrete dynamics of an HMM in both directions. This aspect has significant implications on the expressiveness of the model and its ability to capture the underlying dynamics of many interesting physical applications and avoiding redundancies in the number of regions required to explain the data.

The remainder of this paper will focus on using the proposed hybrid models in two manners: An \textbf{open-loop} setting that treats the control $\vec{u}$ as an exogenous input and is used for automatic system identification of general nonlinear dynamics via decomposition into continuous and discrete switching dynamics. Here we assume all regions to be first-order Markovian linear systems in the state $\vec{x}$. A \textbf{closed-loop} setting, that assumes the controls $\vec{u}$ to originate from a general nonlinear controller that is to be decomposed and inferred simultaneously with the dynamics. In this setting, the controllers can be easily extended to be polynomial in the state $\vec{x}$ or multi-step auto-regressive to incorporate past observations. Such controllers can approximate time-variant policies and lead to overall smoother control, possibly more suitable for the application on real platforms.


\section{Inference of Switching Dynamics and Control}
Different schemes of Bayesian inference approaches have been proposed for learning (r)AR-HMMs \citep{fox2009nonparametric, linderman2017recurrent}, which have relied on computationally costly Gibbs sampling techniques. In this section, we propose an efficient expectation-maximization/Baum-Welch (EM/BW) algorithm \citep{baum1970maximization, dempster1977maximum} to infer the set of parameters $\vec{\theta}$ of an rAR-HMM given multiple time series of observations $\mat{Y}$. Our approach is closely related to the Baum-Welch algorithms proposed in \citep{bengio1995input} and \citep{daniel2016probabilistic}.

\paragraph{Baum-Welch Lower Bound:}Consider again the model of an rAR-HMM in Figure~\ref{fig:rarhmm}, in which the continuous state $\vec{x}$ and action
$\vec{u}$ are observed quantities, while the $K$-region indicators $z$ are hidden. For inferring  the model parameters we generate a dataset $\vec{\mathcal{D}}$
consisting of $N$ trajectories of length $T$ with $\mat{Y}^{n}_{1:T}=\{(\vec{x}_{1}^{n},\vec{u}_{1}^{n}),...,(\vec{x}_{T}^{n},\vec{u}_{T}^{n})\}$
and $\vec{z}^{n}_{1:T}=\{z_{1}^{n},...,z_{T}^{n}\}$. We drop the time indexing to avoid a cluttered notation. The learning objective is to maximize the log-likelihood of the observed data $\vec{\mathcal{D}}$ given the parameters $\vec{\theta} = \{\vec{\pi}, \vec{\mu}_{1,k}, \vec{\Omega}_{1,k}, \mat{\psi}(\vec{\omega}_k), \mat{A}_k, \mat{B}_k, \vec{c}_k, \mat{\Lambda}_k, \mat{K}_k, \mat{\Sigma}_k \} ~\forall k\in[1,K]$
\begin{equation}
	\vec{\theta}_{\text{ML}} = \argmax_{\vec{\theta}} ~ \ln p(\vec{\mathcal{D}} | \vec{\theta})
	= \argmax_{\vec{\theta}} ~ \ln \prod_{n=1}^{N} \sum_{\vec{z}^{n}} p(\mat{Y}^{n},\vec{z}^{n} | \vec{\theta)},
	\label{eq:max_lik}
\end{equation}
where $p(\mat{Y}^{n},\vec{z}^{n}|\vec{\theta})$ is the complete-data likelihood of a single trajectory and factorizes to the following form according to our graphical model
\begin{equation}
	p(\mat{Y}^{n},\vec{z}^{n}|\vec{\theta}) = p(z_{1}^{n}) p(\vec{x}_{1}^{n}|z_{1}^{n}) \prod_{t=1}^{T} p(z_{t+1}^{n}|z_{t}^{n},\vec{x}_{t}^{n},\vec{u}_{t}^{n}) p(\vec{u}_{t}^{n}|\vec{x}_{t}^{n},z_{t}^{n}) p(\vec{x}_{t+1}^{n}|\vec{x}_{t}^{n},\vec{u}_{t}^{n},z_{t+1}^{n}).
	\label{eq:traj_lik}
\end{equation}


The optimization of Equation~\eqref{eq:max_lik} poses a technical difficulty represented by the summation over all possible trajectories of the hidden variables $\vec{z}^{n}$, which is of computational complexity $\mathcal{O}(NK^{T})$ and is intractable in most cases. This issue is overcome by introducing the variational posterior distributions over the hidden variables $q_{\vec{z}^{n}}(\vec{z}^{n})$, and deriving a lower bound on the log-likelihood
\begin{equation}
	\ln\prod_{n=1}^{N} p(\mat{Y}^{n}|\vec{\theta}) = \ln\prod_{n=1}^{N}\sum_{\vec{z}^{n}} p(\mat{Y}^{n},\vec{z}^{n}|\vec{\theta}) \geq \sum_{n=1}^{N}\sum_{\vec{z}^{n}} q_{\vec{z}^{n}}(\vec{z}^{n}) \ln\frac{p(\mat{Y}^{n},\vec{z}^{n}|\vec{\theta})}{q_{\vec{z}^{n}}(\vec{z}^{n})}.
	\label{eq:em_bound}
\end{equation}

We can find a point estimate of the parameters $\vec{\theta}_{\text{ML}}$ by following the standard scheme of EM of alternating between an expectation step (E-step), in which the lower bound in Equation~\eqref{eq:em_bound} is maximized with respect to the variational distributions $q_{\vec{z}^{n}}(\vec{z}^{n})$ given an estimate $\hat{\vec{\theta}}$, and a maximization step (M-step), that updates $\vec{\theta}$ given $q_{\vec{z}^{n}}(\vec{z}^{n})$.

\paragraph{Expectation Step:} By further decomposing Equation~\eqref{eq:em_bound} and isolating $q_{\vec{z}^{n}}(\vec{z}^{n})$ we arrive at
\begin{equation}
	\ln\prod_{n=1}^{N} p(\mat{Y}^{n}|\vec{\theta}) \geq \sum_{n=1}^{N}\ln p(\mat{Y}^{n}|\vec{\theta})-\sum_{n=1}^{N}\text{KL}\left(q_{\vec{z}^{n}}(\vec{z}^{n})||p(\vec{z}^{n}|\mat{Y}^{n},\vec{\theta})\right),
\end{equation}
resulting in the optimal posterior distributions $q_{\vec{z}^{n}}(\vec{z}^{n}) = p(\vec{z}^{n} | \vec{x}_{1:T}^{n}, \vec{u}_{1:T}^{n}, \vec{\theta})$ after which the bound is tight, if the modeled posteriors $q_{\vec{z}^{n}}(\vec{z}^{n})$ lie in the same family of the true distributions. These quantities are the smoothed posterior marginals $p(z_{t}^{n}|\vec{x}_{1:T}^{n}, \vec{u}_{1:T}^{n})$ and can be computed by a forward-backward algorithm \citep{murphy2012}
\begin{equation}
p(z_{t}^{n}=k | \vec{x}_{1:T}^{n}, \vec{u}_{1:T}^{n}) \propto p(z_{t}^{n}=k | \vec{x}_{1:t}^{n}, \vec{u}_{1:t}^{n}) p(\vec{x}_{t+1:T}^{n}, \vec{u}_{t+1:T}^{n}|z_{t}^{n}=k, \vec{x}_{t}^{n}, \vec{u}_{t}^{n}).
\end{equation}

\paragraph{Maximization Step:} By taking the result of the E-step and reformulating Equation~\eqref{eq:em_bound} we can write the lower bound in terms of $\vec{\theta}$ and $\vec{\hat{\theta}}$, where $\vec{\hat{\theta}}$ is the parameter vector from the last M-step
\begin{equation}
	\mathcal{L} =  Q(\vec{\theta},\vec{\hat{\theta}}) = \sum_{n=1}^{N}\sum_{\vec{z}^{n}}p(\vec{z}^{n}|\mat{Y}^{n}, \vec{\hat{\theta}})\ln p(\mat{Y}^{n},\vec{z}^{n}|\vec{\theta}).
\end{equation}
This function is first treated under distribution-normalizing constraints via the method of Lagrangian multipliers before arriving at the complete-log-likelihood that can be maximized w.r.t. $\vec{\theta}$.

\section{Empirical Evaluation}
To analyze the power of auto-regressive models we construct two evaluation scenarios \footnote{Source code can be found on \url{https://github.com/hanyas/sds}}. On the one hand, we want to quantify the quality of open-loop learned models and their ability to capture the underlying dynamics and compare them to state-of-the-art models in \emph{long-horizon} and \emph{small-data} regimes. On the other hand, we use the closed-loop models to learn local switching controllers in an imitation learning setting and replace complicated policy structures.

\subsection{Model Learning and System Identification}
In this experiment, we focus on rAR-HMMs with exogenous inputs. Our aim is to learn the dynamics of three simulated systems; a bouncing ball, an actuation-constrained pendulum, and a cart-pole system. We compare the predictive power of rAR-HMMs to classical AR-HMMs, Feedforward Neural Nets (FNN), Gaussian Processes (GP), Long-Short-Term Memory Networks (LSTM) \citep{hochreiter1997long} and Recurrent Neural Networks (RNN). During evaluation, we collect a training and a test dataset. The training dataset is randomly split into 24 groups each containing a subset of trajectories and used to train different instances of all models, which are then all tested on the test dataset. All neural models have 2 hidden layers, which we test for a variety of different layer sizes, $[16, 32, 64, 128, 256, 512]$ for FNNs, $[16, 32, 64, 128, 256]$ for RNNs, and $[16, 32, 64, 128]$ for LSTMs. In the case of the auto-regressive models (r)AR-HMMs, we test a number of components $K$, dependent on the task. As a comparison metric, we evaluate the Normalized Mean Square Error (NMSE), averaged over the 24 splits, for a range of horizons by combing through the test trajectories step by step and predicting the given horizon. We also briefly compare the model complexity in terms of the total number of parameters of each representation.

\begin{figure}[htb!]
	\begin{adjustbox}{valign=t,minipage={.33\textwidth}}
\begin{tikzpicture}

\definecolor{color0}{rgb}{0.5803921568627451,0.,0.8274509803921568}
\definecolor{color1}{rgb}{0.83921568627451,0.152941176470588,0.156862745098039}
\definecolor{color2}{rgb}{0.12156862745098,0.466666666666667,0.705882352941177}
\definecolor{color3}{rgb}{0.,0.,0.}
\definecolor{color4}{rgb}{0.172549019607843,0.627450980392157,0.172549019607843}
\definecolor{color5}{rgb}{1,0.498039215686275,0.0549019607843137}

\begin{axis}[
grid style={line width=.1pt, draw=gray!10},major grid style={line width=.2pt,draw=gray!50},
minor tick num=3,
width=5.25cm,
height=4cm,
xmin=1, xmax=100,
ymin=0.0, ymax=1.0,
try min ticks=3,
tick align=inside,
tick pos=left,
x grid style={white!69.01960784313725!black},
y grid style={white!69.01960784313725!black},
xtick style={color=black},
ytick style={color=black},
minor tick length = 3pt,
major tick length = 6pt,
xmajorgrids,
ymajorgrids,
xlabel=$h$,
x label style={yshift=.5em},
ylabel=NMSE,
title=Bouncing Ball,
]

\addplot [very thick, color0, forget plot, mark=*]
table {%
	1 0.235030706546711
	20 6.62271668359716
	40 34.771339449703
};

\addplot [very thick, color1, forget plot, mark=*]
table {%
	1 0
	20 1.20274161001059e-16
	40 3.14563190310461e-16
	60 5.44009282066327e-16
	80 6.30976752328631e-16
	100 1.02325555436285e-15
};

\addplot [very thick, color2, forget plot, mark=*]
table {%
	1 0.00030404281925273
	20 0.0532903685266839
	40 0.226063863046134
	60 0.472249145723833
	80 0.711831973238396
	100 1.01760005742054
};

\addplot [very thick, color4, forget plot, mark=*]
table {%
	1 0.000871201859130784
	20 0.0387563169694271
	40 0.142061174714052
	60 0.288511154329654
	80 0.422152936142324
	100 0.767922113441054
};

\addplot [very thick, color5, forget plot, mark=*]
table {%
	1 0.00110803414651805
	20 0.00430101038482412
	40 0.0277253536132656
	60 0.0485589364245783
	80 0.099781158564859
	100 0.152765989523651
};

\end{axis}

\end{tikzpicture}%
	\end{adjustbox}
	\begin{adjustbox}{valign=t,minipage={.33\textwidth}}
		\hspace{0.2cm}
\begin{tikzpicture}

\definecolor{color0}{rgb}{0.5803921568627451,0.,0.8274509803921568}
\definecolor{color1}{rgb}{0.83921568627451,0.152941176470588,0.156862745098039}
\definecolor{color2}{rgb}{0.12156862745098,0.466666666666667,0.705882352941177}
\definecolor{color3}{rgb}{0.,0.,0.}
\definecolor{color4}{rgb}{0.172549019607843,0.627450980392157,0.172549019607843}
\definecolor{color5}{rgb}{1,0.498039215686275,0.0549019607843137}

\begin{axis}[
grid style={line width=.1pt, draw=gray!10},major grid style={line width=.2pt,draw=gray!50},
minor tick num=3,
width=5.25cm,
height=4cm,
xmin=1, xmax=25,
ymin=0.0, ymax=1.0,
try min ticks=3,
tick align=inside,
tick pos=left,
x grid style={white!69.01960784313725!black},
y grid style={white!69.01960784313725!black},
xtick style={color=black},
ytick style={color=black},
minor tick length = 3pt,
major tick length = 6pt,
xmajorgrids,
ymajorgrids,
title=Pendulum (Joint),
]

\addplot [very thick, color0, forget plot, mark=*]
table {%
	1 0.0433942364510538
	5 0.231992998184242
	10 0.537579311458594
	15 0.949035459749998
	20 1.44250281103549
	25 1.96283564211962
};

\addplot [very thick, color1, forget plot, mark=*]
table {%
	1 0.0257828392223513
	5 0.0295060100780364
	10 0.0507961320029683
	15 0.0673253714134016
	20 0.0263532575774912
	25 0.031961852409164
};

\addplot [very thick, color2, forget plot, mark=*]
table {%
	1 0.0382714437242998
	5 0.17788189312194
	10 0.304214714917815
	15 0.472970024829159
	20 0.688004861016135
	25 0.873426973758582
};

\addplot [very thick, color3, forget plot, mark=*]
table {%
	1 0.0361617067888579
	5 0.173274024141512
	10 0.374122434947112
	15 0.630209741194252
	20 0.937563059523567
	25 1.23249582450851
};

\addplot [very thick, color4, forget plot, mark=*]
table {%
	1 0.0882841667298758
	5 0.158903188329079
	10 0.245306465307006
	15 0.363251774312412
	20 0.510458737210679
	25 0.652393590662945
};

\addplot [very thick, color5, forget plot, mark=*]
table {%
	1 0.131693914679793
	5 0.173365171992736
	10 0.173065528038247
	15 0.170793791449654
	20 0.18774091002078
	25 0.228282988244238
};

\end{axis}

\end{tikzpicture}%
	\end{adjustbox}
	\begin{adjustbox}{valign=t,minipage={.33\textwidth}}
\begin{tikzpicture}

\definecolor{color0}{rgb}{0.5803921568627451,0.,0.8274509803921568}
\definecolor{color1}{rgb}{0.83921568627451,0.152941176470588,0.156862745098039}
\definecolor{color2}{rgb}{0.12156862745098,0.466666666666667,0.705882352941177}
\definecolor{color3}{rgb}{0.,0.,0.}
\definecolor{color4}{rgb}{0.172549019607843,0.627450980392157,0.172549019607843}
\definecolor{color5}{rgb}{1,0.498039215686275,0.0549019607843137}

\begin{axis}[
grid style={line width=.1pt, draw=gray!10},major grid style={line width=.2pt,draw=gray!50},
minor tick num=3,
width=5.25cm,
height=4cm,
xmin=1, xmax=25,
ymin=0.0, ymax=1.0,
try min ticks=3,
tick align=inside,
tick pos=left,
x grid style={white!69.01960784313725!black},
y grid style={white!69.01960784313725!black},
xtick style={color=black},
ytick style={color=black},
minor tick length = 3pt,
major tick length = 6pt,
xmajorgrids,
ymajorgrids,
title=Cart-Pole (Joint),
]

\addplot [very thick, color0, forget plot, mark=*]
table {%
	1 0.0227243332690666
	5 0.12263581211881
	10 0.289895464047155
	15 0.602319722210103
	20 1.21786160359631
	25 2.44306231900854
};

\addplot [very thick, color1, forget plot, mark=*]
table {%
	1 0.0222152200378663
	5 0.0463178040205261
	10 0.0530472129505323
	15 0.0799486233848701
	20 0.114470454842913
	25 0.157743538050252
};

\addplot [very thick, color2, forget plot, mark=*]
table {%
	1 0.0259829976321843
	5 0.0947539251573511
	10 0.189066026588177
	15 0.313350482334477
	20 0.484460553696721
	25 0.706225247991407
};

\addplot [very thick, color3, forget plot, mark=*]
table {%
	1 0.0461290182411773
	5 0.166783248341553
	10 0.327936273982853
	15 0.551084030892741
	20 0.858099390627054
	25 1.29216838067192
};

\addplot [very thick, color4, forget plot, mark=*]
table {%
	1 0.0487288227448388
	5 0.125282260892913
	10 0.288311559527632
	15 0.513626573147913
	20 0.798763816236717
	25 1.15127949825997
};

\addplot [very thick, color5, forget plot, mark=*]
table {%
	1 0.125252909008027
	5 0.19514871822825
	10 0.304507907558736
	15 0.455389788250555
	20 0.631458037490151
	25 0.863753395592968
};

\end{axis}

\end{tikzpicture}%
	\end{adjustbox}

	\begin{adjustbox}{valign=t,minipage={.33\textwidth}}
		\vspace{0.75cm}
		\hspace{0.25cm}
\begin{tikzpicture}

\definecolor{color0}{rgb}{0.5803921568627451,0.,0.8274509803921568}
\definecolor{color1}{rgb}{0.83921568627451,0.152941176470588,0.156862745098039}
\definecolor{color2}{rgb}{0.12156862745098,0.466666666666667,0.705882352941177}
\definecolor{color3}{rgb}{0.,0.,0.}
\definecolor{color4}{rgb}{0.172549019607843,0.627450980392157,0.172549019607843}
\definecolor{color5}{rgb}{1,0.498039215686275,0.0549019607843137}

\begin{axis}[
hide axis,
width=5cm,
height=4cm,
xmin=0, xmax=1,
ymin=0, ymax=1.0,
legend cell align={center},
legend columns=2,
legend style={/tikz/every even column/.append style={column sep=0.1}},
]

\addlegendimage{no markers, line width=2pt, color1}
\addlegendentry{\small rAR-HMM};
\addlegendimage{no markers, line width=2pt, color2}
\addlegendentry{\small FNN};
\addlegendimage{no markers, line width=2pt, color0}
\addlegendentry{\small AR-HMM};
\addlegendimage{no markers, line width=2pt, color4}
\addlegendentry{\small RNN};
\addlegendimage{no markers, line width=2pt, color5}
\addlegendentry{\small LSTM};
\addlegendimage{no markers, line width=2pt, color3}
\addlegendentry{\small GP};

\end{axis}

\end{tikzpicture}%
	\end{adjustbox}
	\begin{adjustbox}{valign=t,minipage={.33\textwidth}}
		\vspace{-0.5cm}
		\hspace{-0.35cm}
\begin{tikzpicture}

\definecolor{color0}{rgb}{0.5803921568627451,0.,0.8274509803921568}
\definecolor{color1}{rgb}{0.83921568627451,0.152941176470588,0.156862745098039}
\definecolor{color2}{rgb}{0.12156862745098,0.466666666666667,0.705882352941177}
\definecolor{color3}{rgb}{0.,0.,0.}
\definecolor{color4}{rgb}{0.172549019607843,0.627450980392157,0.172549019607843}
\definecolor{color5}{rgb}{1,0.498039215686275,0.0549019607843137}

\begin{axis}[
grid style={line width=.1pt, draw=gray!10},major grid style={line width=.2pt,draw=gray!50},
minor tick num=3,
width=5.25cm,
height=4cm,
xmin=1, xmax=25,
ymin=0.0, ymax=0.01,
try min ticks=3,
tick align=inside,
tick pos=left,
x grid style={white!69.01960784313725!black},
y grid style={white!69.01960784313725!black},
xtick style={color=black},
ytick style={color=black},
minor tick length = 3pt,
major tick length = 6pt,
xmajorgrids,
ymajorgrids,
scaled y ticks=base 10:2,
ylabel=NMSE,
xlabel=$h$,
x label style={yshift=.5em},
title=Pendulum (Trig),
]

\addplot [very thick, color0, forget plot, mark=*]
table {%
	1 7.10069078216302e-05
	5 0.00107761481926122
	10 0.00476652183408213
	15 0.0139320890850907
	20 0.0356002989938119
	25 0.0848468043787902
};

\addplot [very thick, color1, forget plot, mark=*]
table {%
	1 5.37466144402014e-05
	5 0.000479634561744509
	10 0.00137989587354345
	15 0.00279563341660977
	20 0.00480973411924143
	25 0.00737913226358328
};

\addplot [very thick, color2, forget plot, mark=*]
table {%
	1 5.35421047701619e-05
	5 0.000499223074218273
	10 0.00144089534795381
	15 0.0028619906629223
	20 0.00491442436576992
	25 0.00763891464074248
};

\addplot [very thick, color3, forget plot, mark=*]
table {%
	1 4.8843720774532e-05
	5 0.000443084743591021
	10 0.00135322244188732
	15 0.0028846112505983
	20 0.00516354244817159
	25 0.00809439056113847
};

\addplot [very thick, color4, forget plot, mark=*]
table {%
	1 0.00111676939422515
	5 0.002459243519513
	10 0.00498783785338283
	15 0.00928578769354706
	20 0.0162755268405473
	25 0.0267098120255512
};

\addplot [very thick, color5, forget plot, mark=*]
table {%
	1 0.000399287298273873
	5 0.000924219860866438
	10 0.00165151055053916
	15 0.00263778123493364
	20 0.00394097726906676
	25 0.00540888926220949
};

\end{axis}

\end{tikzpicture}%
	\end{adjustbox}
	\begin{adjustbox}{valign=t,minipage={.33\textwidth}}
		\vspace{-0.5cm}
		\hspace{0.05cm}
\begin{tikzpicture}

\definecolor{color0}{rgb}{0.5803921568627451,0.,0.8274509803921568}
\definecolor{color1}{rgb}{0.83921568627451,0.152941176470588,0.156862745098039}
\definecolor{color2}{rgb}{0.12156862745098,0.466666666666667,0.705882352941177}
\definecolor{color3}{rgb}{0.,0.,0.}
\definecolor{color4}{rgb}{0.172549019607843,0.627450980392157,0.172549019607843}
\definecolor{color5}{rgb}{1,0.498039215686275,0.0549019607843137}

\begin{axis}[
grid style={line width=.1pt,draw=gray!10},major grid style={line width=.2pt,draw=gray!50},
minor tick num=3,
width=5.25cm,
height=4cm,
xmin=1, xmax=25,
ymin=0.0, ymax=0.1,
try min ticks=3,
tick align=inside,
tick pos=left,
x grid style={white!69.01960784313725!black},
y grid style={white!69.01960784313725!black},
xtick style={color=black},
ytick style={color=black},
minor tick length = 3pt,
major tick length = 6pt,
xmajorgrids,
ymajorgrids,
scaled y ticks=base 10:2,
xlabel=$h$,
x label style={yshift=.5em},
title=Cart-Pole (Trig),
]

\addplot [very thick, color0, forget plot, mark=*]
table {%
	1 0.000508253515858238
	5 0.00637410213910227
	10 0.0263480026651941
	15 0.0694938935733579
	20 0.162250155085367
	25 0.363599980040829
};

\addplot [very thick, color1, forget plot, mark=*]
table {%
	1 0.000398128942417908
	5 0.00457149088377579
	10 0.0141101744984037
	15 0.0270536977486792
	20 0.0438685670271435
	25 0.0678419560470744
};

\addplot [very thick, color2, forget plot, mark=*]
table {%
	1 0.000272330229904382
	5 0.00418600022075378
	10 0.0133983573675247
	15 0.0270235642274459
	20 0.0465533118815869
	25 0.07554559473038
};

\addplot [very thick, color3, forget plot, mark=*]
table {%
	1 0.000382622313029227
	5 0.00585130784062151
	10 0.0179102582107519
	15 0.0335194156056009
	20 0.0542072681366596
	25 0.08282618171377
};

\addplot [very thick, color4, forget plot, mark=*]
table {%
	1 0.022775359001483
	5 0.0782490397651167
	10 0.195268718435208
	15 0.385831436837359
	20 0.665696526544842
	25 1.00367594485003
};

\addplot [very thick, color5, forget plot, mark=*]
table {%
	1 0.0109649026534586
	5 0.0300274562394637
	10 0.0592872132115048
	15 0.0994525101364043
	20 0.158471624564303
	25 0.24364870463633
};

\end{axis}

\end{tikzpicture}%
	\end{adjustbox}
	\vspace{-0.35cm}
	\caption{Comparing the $h$-step NMSE of rAR-HMMs to other expressive models, averaged over 24 data splits. Evaluation on three dynamical systems, a bouncing ball, a pendulum and a cart-pole. In this small-data regime scenario, rAR-HMMs exhibit the most consistent approximation capabilities.}
	\label{fig:identification}
	\vspace{-0.15cm}
\end{figure}
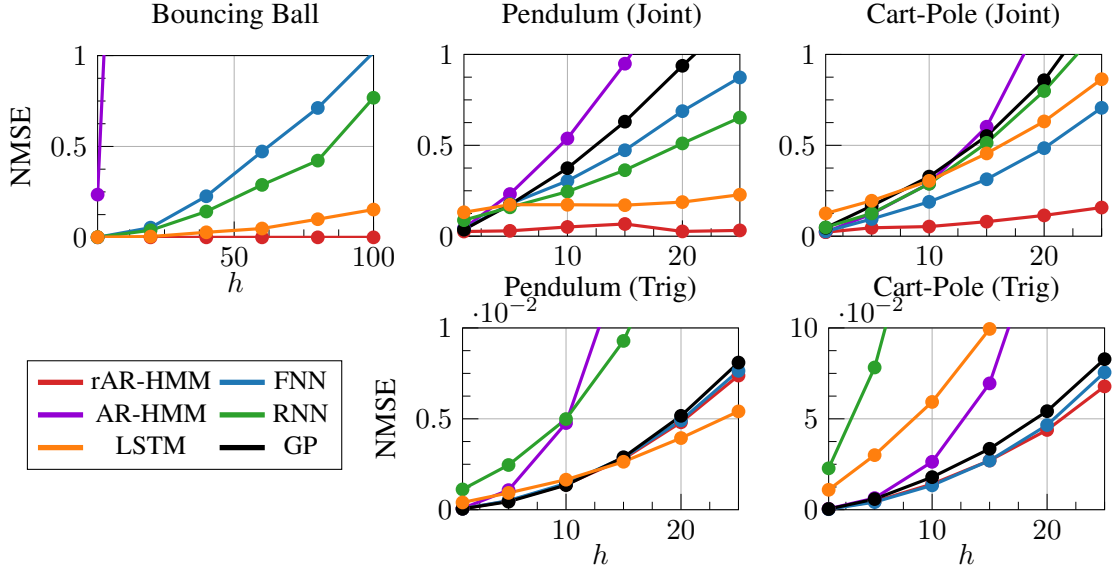

\paragraph{Bouncing Ball} A ball is simulated falling under the influence of gravity until it hits the ground and bounces upwards again. This system is a canonical example of a two-regime two-dimensional hybrid system, because of the hard velocity switch at the moment of impact. We simulate the dynamics with a frequency of $20~\textrm{Hz}$ and collect 25 trajectories for training with different initial heights and velocities, each 30 seconds long. This dataset is split 24 folds with 10 trajectories, $10\times150$ data points in each subset. The test dataset consists of 5 trajectories, each 30 seconds long. We evaluate the NMSE for horizons $h =\{1, 20, 40, 60, 80\}$ time steps. We did not evaluate a GP model in this setting, due to the long prediction horizons that result in a very high computation effort during evaluation time. The (r)AR-HMM models are tested for $K=2$. The logistic link function of the rAR-HMM is a neural net with one hidden layer containing 16 neurons. The results depicted in Figure~\ref{fig:identification} show that the rAR-HMM model is able to approximate the true dynamics very well and outperforms the standard AR-HMM and all other neural models.

\begin{table}[h!]
	\centering
	\begin{tabular}{ c | c | c | c | c | c }
		\hline
	               & Bouncing      & Pend. (Joint) & Pend. (Trig)  & Cart. (Joint) & Cart. (Trig)  \\
		\hline
		AR-HMM     &     22 (2)    &    180 (9)    &     130 (5)   &    287 (7)    &   275 (5)     \\
		\hline
		rAR-HMM    &     86 (2)    &    468 (9)    &     582 (9)   &    575 (7)    &   711 (7)     \\
		\hline
		FNN        &   1250 (32)   &   546 (64)    &   1315 (32)   &   1380 (32)   &   1445 (32)   \\
		\hline
		RNN        &  12866 (64)   &  50306 (128)  &   3427 (32)   &  50820 (128)  &  51077 (128)  \\
		\hline
		LSTM       & 200450 (128)  &  51074 (64)   &  51395 (64)   & 201732 (128)  & 202373 (128)  \\
		\hline

	\end{tabular}
	\caption{Qualitative comparison of model complexity of the best performing representations in Figure~\ref{fig:identification}. The values reflect the total number of parameters of each model. The values in parentheses reflect the size of the hidden layers of the neural models and the number of discrete components of the auto-regressive models, respectively.}
	\vspace{-0.25cm}
\end{table}

\paragraph{Pendulum and Cart-Pole} These two systems are classical benchmarks from the control literature. Here we consider two different observation models, one in the wrapped joint space, which includes a sharp discontinuity in the pendulum/pole angles, and a second model in a smooth trigonometric space of the angles.
Both dynamics are simulated with a frequency of $100~\textrm{Hz}$. We collect 25 training trajectories starting from different initial conditions and applying random explorative actions. Each trajectory is 2.5 seconds long. The 24-splits each consist of 10 trajectories, $10\times250$ data points. The test dataset consists of 5 trajectories, each 2.5 seconds long. Forecasting accuracy is evaluated for horizons  $h =\{1, 5, 10, 15, 20, 25\}$. The (r)AR-HMM models are tested for $K=\{3, 5, 7, 9\}$ on both tasks. The logistic link function of the rAR-HMM, in both tasks, is a neural net with one hidden layer containing 24 neurons. The forecast evaluation of both tasks as shown in Figure~\ref{fig:identification} provides empirical evidence for the representation power of rAR-HMMs in both smooth and discontinuous state spaces. FNNs and GPs perform equally well in the smooth trigonometric space of observations, struggle however in the discontinuous space, similar to RNNs and LSTMs.

\begin{figure}[t!]
	\begin{minipage}[t]{0.33\columnwidth}%
		\includegraphics[page=1]{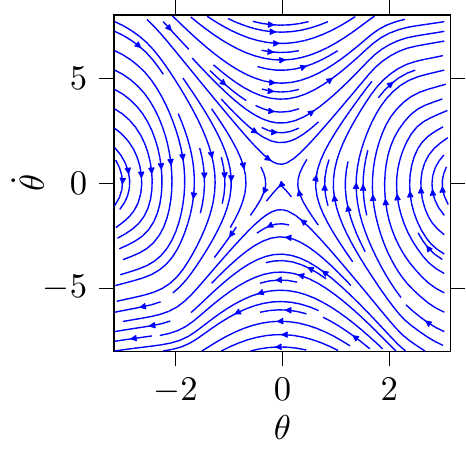}
	\end{minipage}\hspace*{.5cm}%
	\begin{minipage}[t]{0.33\columnwidth}%
		\includegraphics[page=1]{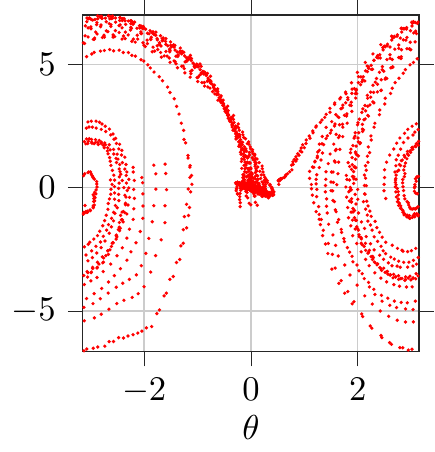}
	\end{minipage}\hspace*{0.cm}%
	\begin{minipage}[t]{0.33\columnwidth}%
		\includegraphics[page=1]{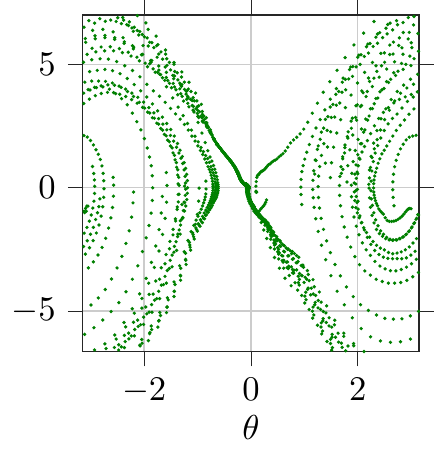}
	\end{minipage}
	\vspace{-0.75cm}
	\caption{Phase portraits of the pendulum experiments. The identified unforced pendulum dynamics is in blue (left), while the learned stationary hybrid policy by imitation of time-variant Optimal Control (OC) trajectories is depicted in red (middle) and the stationary hybrid policy learned by imitation of a global Soft Actor-Critic (SAC) policy is seen in green (right).}
	\label{fig:pendulum_phase}
\end{figure}

\begin{figure}[t!]
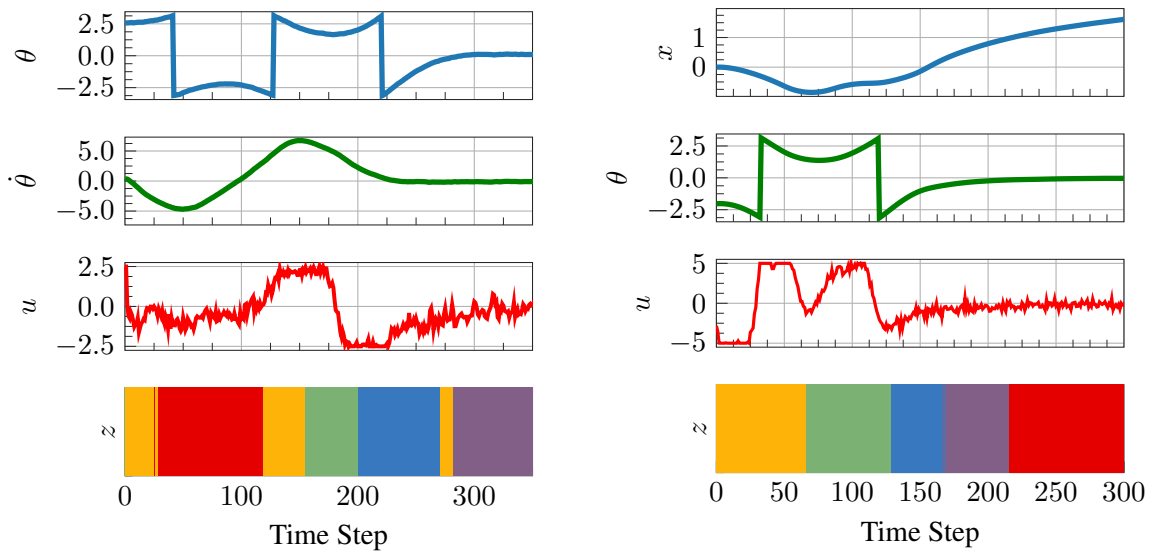

	\begin{minipage}{0.45\columnwidth}
		\input{figures/gps_pendulum_imitation_example.tex}
	\end{minipage}\hspace{1.cm}
	\begin{minipage}{0.45\columnwidth}
		\input{figures/sac_cartpole_imitation_example.tex}
	\end{minipage}
	\vspace{-0.75cm}
	\caption{Sample trajectories from the learned stationary hybrid policies on the pendulum (left) and cart-pole (right) environments. Both policies are able to consistently solve both tasks, while relying on simple local representations of the feedback controllers.}
	\vspace{-0.5cm}
	\label{fig:trajectory_examples}
\end{figure}

\paragraph{Policy Distillation:}
For this evaluation, we consider the full rAR-HMM with endogenous inputs, that stem from an internal policy whose parameters are to be inferred. For this purpose, we supply policies from two different experts for the pendulum and cart-pole environments. One policy is a time-variant optimal controller computed via iterative Differential Dynamic Programming \citep{jacobson1970differential} and the other is a reinforcement learning agent trained by Soft Actor-Critic (SAC) \citep{haarnoja2018soft}, with 4545 parameters for pendulum and 17537 parameters for cart-pole. For imitation, we chose 1-step auto-regressive hybrid policies with 5 regimes each. The resulting controllers were able to complete the task of swinging up and stabilization of both systems with over 95$\%$ success rate. Figure~\ref{fig:pendulum_phase} shows the phase portrait of the open-loop identified dynamics and closed-loop control during imitation.  Figure~\ref{fig:trajectory_examples} depicts sampled trajectories of the hybrid policies applied to the environments highlighting the switching behavior.

\clearpage 

\section{Related Work}
\label{related}
Hybrid systems have been extensively studied in the control community and are widely used in real-world/real-time applications \citep{borrelli2006mpc, menchinelli2008hybrid}. For an overview on system identification of hybrid systems, we refer the reader to \citep{paoletti2007}, in which the authors introduce a taxonomy of hybrid systems with linear separation boundaries, and review different procedures for clustering and identifying sub-regimes of dynamics, varying from mixed-integer optimization \citep{bemporad2001identification} to Bayesian methods \citep{juloski2005bayesian}.

Hybrid representations also play a central role in the domain of data-driven, general-purpose process modeling and state estimation \citep{ackerson1970state, hamilton1990analysis}. Moreover, different classes of stochastic hybrid systems serve as powerful generative models for complex dynamical behaviors \citep{pavlovic2001learning, mesot2007switching, oh2005data}. The analysis of these models has been traditionally considered under the paradigms of message passing and Markov chain Monte Carlo \citep{fox2009nonparametric, linderman2017recurrent}, however, the recent rise of variational auto-encoders \citep{kingma2013auto} has enabled a new and powerful view on possible inference techniques \citep{becker2019switching}. A distinct property of approaches that rely on auto-encoders is the need to relax the discrete variables in order to do inference.

Switching systems have also served as a powerful tool in various imitation learning approaches. \cite{calinon2010learning} combine traditional HMMs with Gaussian mixture regression to represent trajectory distributions, while \cite{daniel2016probabilistic} use a hidden semi-Markov model to learn hierarchical policies and \cite{burke2019hybrid} introduced switching density networks for system identification and behavioral cloning. Finally, excellent work on hierarchical decomposition of policies in a fully Bayesian framework is introduced by \cite{vsovsic2017bayesian}, albeit under known transition dynamics.

\section{Conclusion}
We have presented a data-driven view for the hierarchical decomposition of nonlinear dynamics and control into simpler units based on the paradigms of hybrid systems and probabilistic graphical models. We have discussed the advantages of such models and demonstrated on a range of classical benchmarks that auto-regressive locally linear models are powerful representations that can capture long-horizon nonlinear dynamics in scenarios with limited data. Furthermore, we have shown that the extension to support endogenous inputs is useful in an imitation learning setup, and can drastically compress over-parameterized neural polices. Moreover, we have used a natural extension of this framework to represent and learn non-Markovian auto-regressive policies.

We view this work as a first step towards a more structured application of machine learning principles in learning-for-control, hoping that by applying regularization through simplicity (Occam's razor), such structures may have a positive effect on the learning procedures that have recently drifted towards ever larger models. Future considerations will focus on more efficient learning algorithms of these graphical models and pursue to highlight their main advantages for the domain of learning-for-control, by extending them to become Bayesian and applying them to environments with discrete events and state-dependent noise.

\acks{This work has received funding from the European Union’s Horizon 2020 research and innovation program under grant agreement \#640554 (SKILLS4ROBOTS).}

\bibliography{references}

\end{document}